%% file: acl_latex.tex
\pdfoutput=1
\documentclass[11pt]{article}
\usepackage[dvipsnames]{xcolor}
\usepackage[final]{acl}

\usepackage{times}
\usepackage{latexsym}

\usepackage[T1]{fontenc}
\usepackage[utf8]{inputenc}

\usepackage{microtype}

\usepackage{inconsolata}

\usepackage{graphicx}
\usepackage{algorithm}
\usepackage{algpseudocodex}
\usepackage{hyperref}
\usepackage{url}
\usepackage[utf8]{inputenc}
\usepackage[T1]{fontenc}    
\usepackage{booktabs}       
\usepackage{amsfonts}       
\usepackage{nicefrac}       
\usepackage{microtype}      
\usepackage{bbm}
\usepackage{tabularx}
\usepackage{graphicx}
\usepackage{multirow}
\usepackage{colortbl}
\usepackage{wrapfig,lipsum}
\usepackage{makecell}
\usepackage{amsmath}
\usepackage{subfigure}
\usepackage[capitalize]{cleveref} 

\definecolor{denim}{rgb}{0.08, 0.38, 0.74}
\usepackage{enumitem}

\algrenewcommand\algorithmicrequire{\textbf{Input:}}
\algrenewcommand\algorithmicensure{\textbf{Output:}}
\algnewcommand{\IfThen}[2]{% \IfThenElse{<if>}{<then>}
  \State \algorithmicif\ #1\ \algorithmicthen\ #2}
\definecolor{comm}{gray}{0.6}
\newcommand{\method}[0]{\textsc{MAgICoRe}}
\newcommand{\acron}[1]{\underline{\textbf{#1}}}
\newcommand{\llama}[0]{\texttt{Llama3-8B-Instruct}}
\newcommand{\gpt}[0]{\texttt{GPT-3.5-Turbo}}

\newcommand{\plus}[1]{{\footnotesize\color{ForestGreen}(+#1\%)}}

\usepackage[framemethod=TikZ]{mdframed}
\definecolor{OurColor}{HTML}{36aa70}
\definecolor{UserExampleBg}{HTML}{ffffff}
\definecolor{UserExampleTitle}{HTML}{545f7f}
\newmdenv[
    roundcorner=5pt,
    backgroundcolor=UserExampleBg,
    linecolor=UserExampleTitle,
    outerlinewidth=0.5pt,
    frametitlebackgroundcolor=UserExampleTitle,
    frametitlefont={\bfseries\color{white}}
]{user_example}

\AtBeginEnvironment{user_example}{\setlength{\parindent}{0pt}}
\title{\textsc{MAgICoRe}: Multi-Agent, Iterative, Coarse-to-Fine\\Refinement for Reasoning}  

\author{
 \textbf{Justin Chih-Yao Chen},
 \textbf{Archiki Prasad},
 \textbf{Swarnadeep Saha},
\\
 \textbf{Elias Stengel-Eskin},
 \textbf{Mohit Bansal}
\\
\\
 UNC Chapel Hill
\\
}

\begin{document}
\maketitle
\begin{abstract}
Large language model (LLM) reasoning can be improved by scaling test-time compute with aggregation, i.e., generating multiple samples and aggregating over them.
While improving performance, this strategy often reaches a saturation point beyond which additional compute provides no return. 
Refinement offers an alternative by using model-generated feedback to improve answer quality. 
However, refinement faces three key challenges:
\textbf{(1) Excessive refinement:} Uniformly refining all instances can cause over-correction and reduce overall performance.
\textbf{(2) Inability to localize and address errors:} LLMs struggle to identify and correct their own mistakes.
\textbf{(3) Insufficient refinement:} Stopping refinement too soon could leave errors unaddressed.
To tackle these issues, we propose \method{}, a framework for \acron{M}ulti-\acron{Ag}ent \acron{I}teration for \acron{Co}arse-to-fine \acron{Re}finement.
\method{} mitigates excessive refinement by categorizing problems as easy or hard, solving easy problems with coarse-grained aggregation, and solving the hard ones with fine-grained multi-agent refinement. 
To better localize errors, we incorporate external step-wise reward model scores, and to ensure sufficient refinement, we iteratively refine the solutions using a multi-agent setup.
We evaluate \method{} on Llama-3-8B and GPT-3.5 and show its effectiveness across seven reasoning datasets.
One iteration of \method{} beats Self-Consistency by $3.4\%$, Best-of-$k$ by $3.2\%$, and Self-Refine by $4.0\%$ even when these baselines use $k=120$, and \method{} uses less than $50\%$ of the compute.
\footnote{Code: \url{https://github.com/dinobby/MAgICoRe}}
\end{abstract}

\begin{figure*}[t!]
    \centering
    \includegraphics[width=\textwidth]{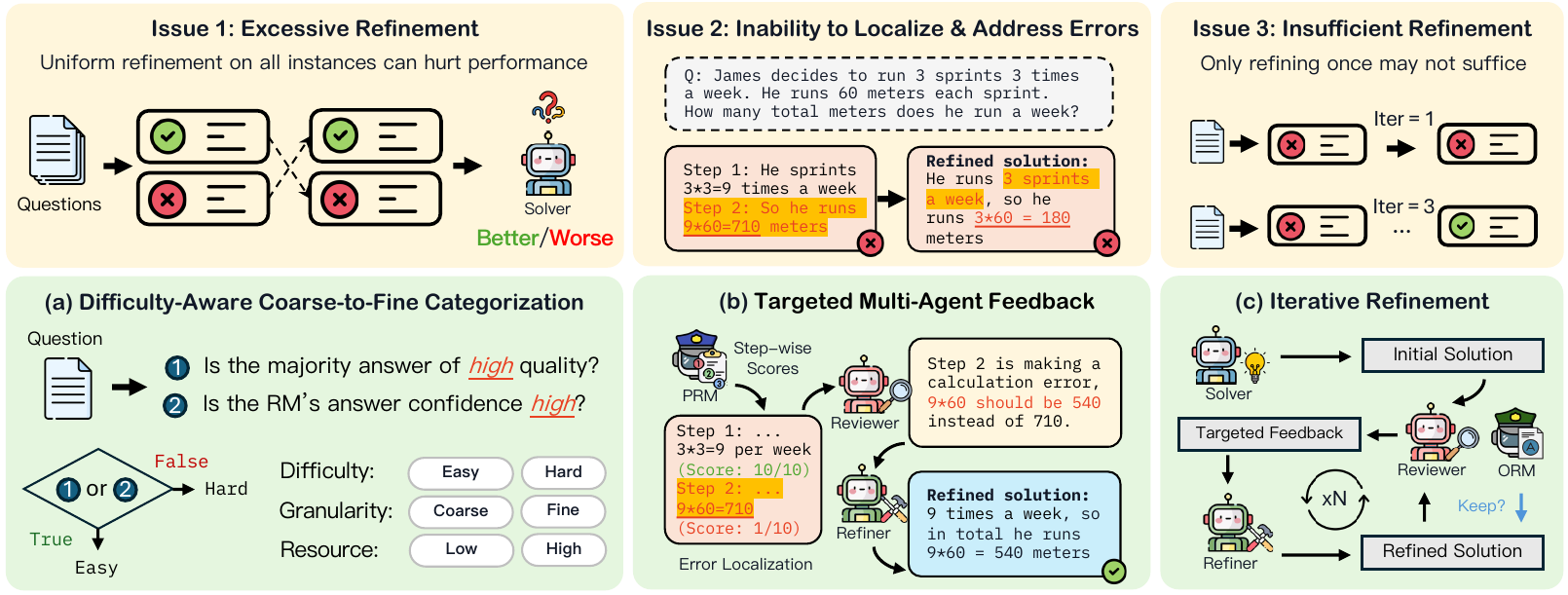}
    \vspace{-1.5em}
    \caption{\textbf{Top:} Three main issues in refinement: 
    \emph{(1) Excessive refinement;} 
    \emph{(2) Inability to localize and address errors;} 
    \emph{(3) Insufficient refinement.} 
    \textbf{Bottom:} Our joint solution to address these issues. \method{} adapts resource allocation based on problem difficulty and refines only when encountering hard problems to avoid excessive refinement. For hard cases requiring refinement, we employ a multi-agent setup that iteratively reviews and refines the solutions based on targeted feedback generated with step-wise PRM scores. 
    }
    \label{fig:intro}
    \vspace{-0.5em}
\end{figure*}

\section{Introduction}
\label{sec:intro}

Imagine a person taking a math exam with problems of varying difficulty; even before answering any question, an effective exam-taker might first distinguish between easier and more challenging problems, 
deciding how much effort to budget for each one \citep{son2000metacognitive, dodeen2015teaching}.
To maximize their score, the student would likely spend minimal time on the easy problems and focus more on the harder ones, refining their answers where needed. 
Misallocating effort could not only waste time but even lower their score, as overthinking simple problems might lead to mistakes; 
similarly failing to dedicate enough thought to hard problems will lead to poor results. 
For Large Language Models (LLMs) performing reasoning tasks, several test-time approaches dedicate more computation to improve performance. 
These approaches sample multiple solutions to the same question and aggregate over the resulting answers (e.g. Self-Consistency \citep[SC;][]{wang2022self}, Best-of-$k$ sampling \citep{lightman2023let,sun2024easytohardgeneralizationscalablealignment,wang2023math}).
However, applying aggregation strategies uniformly may waste computation on easier problems where the performance saturates quickly, and has diminishing gains on the harder problems even when more samples are generated. 
Refinement -- where solutions are instead critiqued and improved upon during resampling -- offers a possible avenue for breaking out of the aggregation rut. 
This mirrors human reasoning, where incorporating feedback (rather than simply retrying) can improve answers, often in an iterative fashion.
For example, a teacher might improve a student's understanding by providing multiple rounds of feedback on a test \citep{pan2021pretest,henry2006test,ken2013feedback}.

While refinement seems promising, it faces three key challenges that current work has yet to fully address, as outlined in \cref{fig:intro}:
\textbf{(1) Excessive refinement}: the LLM must know when to refine and when not to. While refinement can help on incorrectly solved problems, uniformly refining all instances can cause over-refinement, where solutions that were already correct before refinement are ``overthought'' and flipped to incorrect, reducing the overall performance \citep{huang2024large, shridhar2024art, stechly2024selfverification}.
\textbf{(2) Inability to localize and address errors}: LLMs struggle to identify their own mistakes (i.e., steps needing refinement) and struggle to correct them without explicit instructions.
\textbf{(3) Insufficient Refinement}: deciding how much refinement is needed is non-trivial -- stopping refinement early could leave errors unaddressed, i.e., hard problems might be ``underthought'' by a single refinement iteration.

To enable better test-time scaling for aggregation and to address the three issues in refinement, we propose a unified solution, \method{}: \acron{M}ulti-\acron{Ag}ent \acron{I}teration for \acron{Co}arse-to-fine \acron{Re}finement.
As shown in \cref{fig:intro}, our approach leverages external global and local Reward Models (RMs) that enhance both aggregation and refinement processes. 
To avoid excessive refinement, we perform \emph{selective refinement} (see \cref{fig:intro}(a)): 
we start by generating multiple reasoning chains from the LLM and scoring them with the RMs, using the entropy of the final answer distribution to classify examples as easy or hard.
Given LLMs' general inability to localize errors \citep{tyen2024correct}, we leverage \emph{step-by-step scores from a process reward model (PRM)} to help the LLM pinpoint low-scoring steps (which are likely to be incorrect); this process is shown in \cref{fig:intro}(b).
Moreover, to help LLMs refine effectively once the errors have been localized, we propose a \emph{multi-agent system} consisting of three agents: the Solver, the Reviewer, and the Refiner. 
For each problem, the Solver generates reasoning chains, the Reviewer gives targeted feedback based on step-by-step RM scores, and the Refiner improves the solutions using this feedback. 
Finally, to address the issue of insufficient refinement, we \emph{iterate the review-refine process}, using the quality and the entropy of the answers at each iteration as a stopping criterion (cf. \cref{fig:intro}(c)). While these three issues -- selective refinement, error localization, and iterative refinement -- might seem independent, addressing them jointly is more effective. Empirically, \method{} consistently outperforms baselines that tackle these issues in isolation, as confirmed by our ablation studies in \cref{tab:issue_ablation}.

We evaluate \method{} on seven reasoning datasets (including math, commonsense and logical reasoning) with two LLMs: Llama-3-8B and GPT-3.5. Notably, \method{} shows consistent improvements over all aggregation and refinement baselines across datasets and models. 
Specifically, just one iteration of \method{} on Llama-3-8B already outperforms Best-of-$k$ sampling \citep{lightman2023let} by $3.2\%$ and Self-Consistency \citep{wang2022self} by $3.4\%$, while using roughly half of the test-time compute.
\method{} also outperforms a combination of Self-Refine \citep{madaan2023self} and Self-Consistency by $4.0\%$ and these trends also hold true for GPT-3.5.
Moreover, \method{} effectively decides when to use refinement and when \emph{not} to, leading to a $6.4\%$ improvement over the strongest Best-of-$k$ baseline on MATH \citep{hendrycksmath2021}, whereas uniformly applying refinement to all samples can result in a \emph{drop} of $5.2\%$, highlighting the key role played by selective refinement.
\method{} also scales better with more iterations of refinement, scales to stronger base models and RMs, applies to both math/reasoning and commonsense tasks, and continues to improve while the baselines stagnate. 

\begin{figure*}
\vspace{-8pt}
    \centering
    \includegraphics[width=0.99\textwidth]{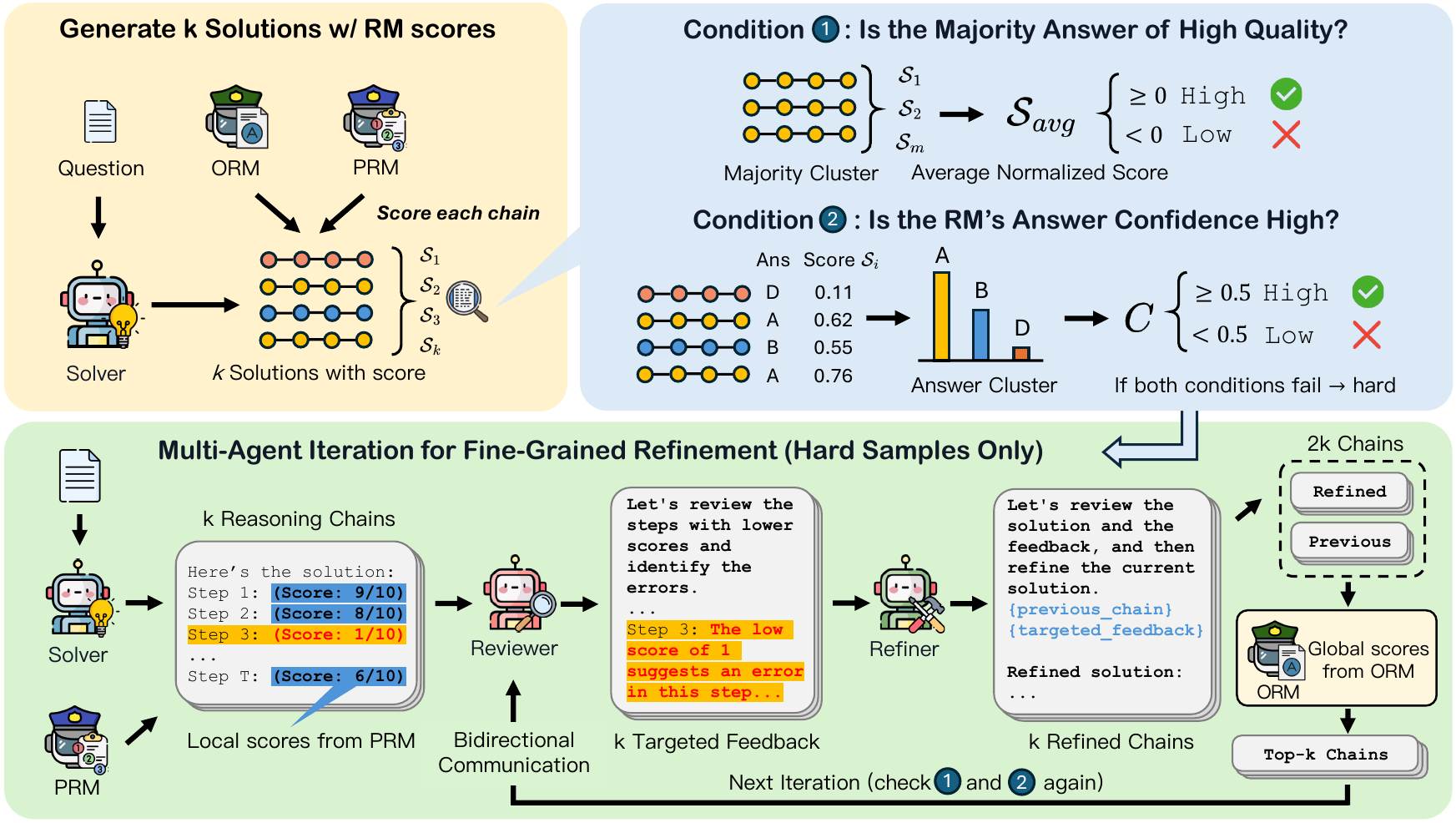}
    \vspace{-0.5em}
    \caption{
    \method{} classify problem difficulty based on two conditions: (1) the quality of the majority answer and (2) the RM's answer confidence. 
    A problem is considered hard when the majority answer receives a low average RM score \emph{and} the answer distribution is flat (i.e., low confidence). 
    For these hard problems, we employ a multi-agent setup --
    The Solver generates $k$ reasoning chains, passing them to the PRM to pinpoint \setlength{\fboxsep}{1pt}\colorbox{Dandelion}{\textcolor{red}{errors}}. The Reviewer turns scores into targeted feedback, and the Refiner improves the $k$ solutions using the Reviewer's feedback.
    This review-refine process repeats until either of the two conditions passes, or a maximum iteration is reached.}
    \label{fig:model}
\end{figure*}

\section{Methodology}
\label{sec:method}
\label{sec:background}
In \method{}, we incorporate three types of models: (1) an LLM interchangeably performing three roles: the Solver, the Reviewer, and the Refiner, (2) an Outcome Reward Model (ORM) for generating global, solution-level correctness score, and (3) a Process Reward Model (PRM) for generating local step-by-step correctness scores. 
Both the ORM and PRM contribute to (1) assessing problem difficulty and (2) final answer selection via Weighted Self-Consistency \citep{li2023making} (see \cref{ssec:wsc}).

\paragraph{Overview.} We present \method{} in \cref{fig:model}. The process begins with the Solver generating $k$ reasoning chains for each problem, followed by the ORM and PRM providing solution-level scores. Next, the input problem's difficulty is classified based on two criteria (top-right of \cref{fig:model}): (1) the quality of the majority answer and (2) the RMs' answer confidence. Refinement is initiated only when the problem is deemed difficult, which occurs when the majority answer receives a low average RM score \emph{and} the answer distribution is flat -- indicating no single answer is significantly better than the others (i.e., low confidence). For these hard samples requiring refinement, we employ a multi-agent setup with three agents: the Solver, the Reviewer, and the Refiner (bottom of \cref{fig:model}). The Reviewer uses the step-wise scores from the PRM to generate targeted feedback, and the Refiner then enhances the $k$ solutions based on this feedback. The review-and-refine cycle can iterate multiple times to ensure sufficient and effective refinement. 

\subsection{Classifying Problem Difficulty}
\label{sec:stage1}
We categorize each problem's difficulty as easy or hard using the following conditions (cf. \cref{fig:model}).\vspace{1pt}\\
\textbf{a) Is the Majority Answer of High Quality?}
The Solver generates $k$ solutions for the input question and we group them by their final answers. From the largest cluster of solutions, we calculate the average RM score and normalize it by the average score across all solutions, denoted as $\mathcal{S}_{avg}$. 
If $\mathcal{S}_{avg} \ge 0$ after normalization, this condition will be \emph{true}, meaning the majority answer is already high-quality (as measured by both ORM and PRM scores, see \cref{sec:cond}), and hence no refinement is needed.
Otherwise, we deem the example to be a possible candidate for refinement and evaluate the second condition below. 
\vspace{2pt}\\\textbf{b) Is Reward Models' Answer Confidence High?}
In this condition, we check if the RMs are confident in any single answer;
if this is not the case, the problem is a possible candidate for refinement. 
We measure confidence via the entropy of the distribution over answers, obtained by weighting answer clusters by their average RM scores, in line with Weighted Self-Consistency \citep{li2023making}. 

\paragraph{Coarse-to-Fine Decision.} If \emph{either} of the conditions is met (the quality of the answer is high \emph{or} the RMs are confident on an answer), an instance is marked as easy and delegated to the coarse-grained method: Weighted Self-Consistency \citep{li2023making}, using the sum of the solution-level scores generated by both ORM and PRM. 
Conversely, if \emph{both} conditions are not satisfied, the instance is marked as hard and delegated to the fine-grained method (as described in \cref{ssec:multiagent}), addressing Issue 1 (excessive refinement) by only refining solutions for the hard problems. 

\subsection{Fine-Grained Multi-Agent Refinement}
\label{ssec:multiagent}
For hard instances that fail both conditions, we need to employ refinement to unlock improvements (see the bottom part of \cref{fig:model}). 
Our refinement setup has three agents: (1) the \textit{Solver}, which generates the initial solution (2) the \textit{Reviewer}, which takes step-wise PRM scores and a reasoning chain as input, and generates targeted feedback that pinpoints the errors within the chain, and (3) the \textit{Refiner}, which takes the feedback generated by the Reviewer to refine the previous chain.
\vspace{2pt}\\\textbf{Solver generates $k$ solutions.} 
The Solver is responsible for generating the initial $k$ solutions.
Recall that in \cref{sec:stage1}, we assess problem difficulty using $k$ generated solutions. 
When a problem is classified as easy, we aggregate the $k$ solution without refinement.
When a problem is classified as hard, we can directly re-use the $k$ solutions \textit{already generated} by the Solver.
\vspace{2pt}\\\textbf{Reviewer generates targeted feedback.} 
To assist the Reviewer in generating useful feedback to localize errors better (``Issue 2'' in \cref{fig:intro}), we supply the Reviewer with external step-wise PRM scores for each step of the solution. The goal of the Reviewer is to incorporate these step-wise correctness scores to generate actionable feedback.
We append these scores to the end of each step and pass the result to the Reviewer. That is, it takes a chain with the PRM scores as input, and is prompted to identify problematic steps that need refinement and possible ways to fix them. 
\vspace{2pt}\\\textbf{Refiner improves solutions w/ feedback.}
Inspired by the finding that LLM can resolve errors when explicitly pointed out \citep{tyen2024correct}, the Refiner agent focuses exclusively on \emph{how the step should be modified} so as to resolve the error based on feedback from the Reviewer.
That is, the Refiner uses the targeted feedback generated by the Reviewer to refine the reasoning chain that was generated by the Solver. 
The prompts for the Reviewer and the Refiner are shown in \cref{sec:prompt}.
\vspace{2pt}\\\textbf{Iterating the refinement process.} 
For some hard instances, one round of refinement may be insufficient, as the Reviewer may have generated some irrelevant feedback or the Refiner may not have fixed the highlighted step adequately (``Issue 3'' in \cref{fig:intro}). 
Thus, the Reviewer and the Refiner need to collaborate with each other over the course of multiple refinement iterations. 
To prevent excessive refinement, we re-evaluate the two conditions described in \cref{sec:stage1} in each iteration. 
The refinement continues until (a) one of the conditions is met, or (b) a predefined maximum number of iterations is reached. 
\vspace{2pt}\\\textbf{Final answer selection.}
The refinement process described above operates on all $k$ chains simultaneously, producing $k$ refined chains in each iteration.
At the end of each iteration, we use the ORM to assess whether the refined solution has improved based on its global correctness score. 
In other words, by the end of each iteration, we have $2k$ reasoning chains -- $k$ initial and $k$ refined -- but retain only the top $k$ based on their global ORM scores. 
Here we choose to base the decision on the ORM score because the targeted feedback is generated with PRM's step-wise scores, so selecting the solution via another scoring model avoids overfitting.
Finally, the answer is selected using Weighted Self-Consistency over these retained top $k$ chains, at the end of each iteration.

\section{Experimental Setup}
\label{sec:expt}
\textbf{Implementation Details.} We develop \method{} with \llama~\citep{llama3modelcard} and \gpt~\citep{openai2023gpt3.5} as the base LLMs.
Based on their strong performance on standard reward modeling evaluations \citep{lambert2024rewardbench}, we choose InternLM-7B \citep{cai2024internlm2} as the ORM, and Math-Shepherd-7B \citep{wang2023math} as the PRM for computing the RM scores. 
By default, we sample $k=40$ reasoning chains in each iteration for \method{}, and the decoding temperature is set to $0.8$. 
The maximum number of iterations is set to $3$, with additional analysis in \cref{fig:iteration} and \cref{tab:more}; we find that after 3, performance saturates, leading us to choose 3 given budget considerations. 
We compare against different categories of strong baselines as follows, and leave more comparisons against prompting-based baselines to \cref{tab:additional_baselines} in the Appendix.

\begin{itemize}[topsep=0pt, wide=1pt, itemsep=0pt, leftmargin=*, after=\strut]
    \item \textbf{Vanilla Prompting.} The first baseline we compare to is zero-shot Chain-of-Thought~\citep{wei2022chain}; note that this only generates one reasoning chain per question without aggregation.
    \item \textbf{Iterative Prompting.} We also compare \method{} to an iterative prompting method, \textbf{Self-Refine (SR)} \citep{madaan2023self}, which refines the initial CoT answer via iteratively prompting the LLM to generate feedback and refine the previous output accordingly.
    \item \textbf{Aggregation-based Methods.} The third category generates multiple samples for each question. 
    Here, we sample $k$ solutions from the same LLM and select the final answer either via $k$-way \textbf{Self-Consistency (SC)} \citep{wang2022self} or according to the highest ORM score (\textbf{Best-of-$k$}) \citep{lightman2023let,sun2024easytohardgeneralizationscalablealignment,wang2023math}. 
    Note that we give these baselines more samples than \method{}.
    \item \textbf{Iterative Baseline with Aggregation.}
    To enable a fair comparison, we also report a stronger version of self-refine by combining \textbf{Self-Refine and Self-Consistency (SR+SC)}, i.e., a baseline that is iterative, refines, and aggregates. Specifically, this baseline applies Self-Refine for $k$ samples in parallel, and the final answer is derived by aggregating the $k$ refined solutions.
\end{itemize}

\noindent\textbf{Datasets.} We evaluate \method{} mainly on five math reasoning datasets. Later in \cref{sec:generalization}, we further show \method{}'s effectiveness on commonsense (ARC-challenge; \citet{clark2018think}) and logical reasoning (Date Understanding; \citet{srivastava2022beyond}) tasks.
The first class of math datasets is math word problems: \textbf{GSM8K}~\citep{cobbe2021training}, \textbf{SVAMP}~\citep{patel2021nlp}, and \textbf{MATH}~\citep{hendrycksmath2021}. GSM8K and SVAMP consist of grade school-level math problems, with 1,312 and 1,000 test samples. 
MATH comprises high-school math competitions spanning diverse topics and a total of 5,000 problems. 
Following previous works \citep{lightman2023let,wang2023math}, we evaluate MATH performance on a representative subset of 500 samples.
We also evaluate on math splits of general benchmarks that test language models' world knowledge and problem-solving abilities over various subjects such as \textbf{MMLU-Math}~\citep{hendryckstest2021, yue2024mammoth} and \textbf{SAT}~\citep{zhong2023agieval} with 974 and 220 test instances respectively.

\section{Results and Analysis}
\begin{table*}[t]
\centering
\resizebox{\textwidth}{!}{%
\begin{tabular}{lllllll} 
\toprule
& \textbf{MMLU} & \textbf{MATH} & \textbf{SVAMP} & \textbf{GSM8K} & \textbf{SAT} & \textbf{Avg.} \\
\midrule
\multicolumn{7}{c}{\textbf{\llama}}\\
\midrule
Zero-shot CoT & 50.4 & 24.2 & 72.4 & 80.1 & 58.2 & 57.1 \\
Self-Refine (Best Iter) & 49.8 & 24.0 & 72.6 & 79.6 & 59.6 & 57.1 \\
\rowcolor{blue!7}Best-of-$k$ ($k = 120$) &  62.6 & 41.4 & 88.7 & 90.1 & 72.4 & 71.0 \\
$k$-way SC ($k = 120$) & 63.0 & 40.6 & 89.8 & 90.3 & 70.5 & 70.8 \\
Self-Refine + $k$-way SC (Best Iter) & 62.3 & 41.0 & 89.2 & 90.3 & 68.0 & 70.2 \\\midrule
\method{} (Iter=1) & 67.3 & 46.0 & \textbf{91.4} & 91.1 & 75.0 & 74.2 \\
\method{} (Iter=2) & 68.4 & 47.2 & 91.1 & \textbf{92.3} & 76.4 & 75.1 \\
\method{} (Iter=3) & \textbf{68.9} \plus{5.6} & \textbf{47.8} \plus{5.2} & 91.3 \plus{1.7} & 91.6 \plus{1.3} & \textbf{78.2} \plus{5.8} & \textbf{75.6} \plus{4.3} \\
\midrule
\multicolumn{7}{c}{\textbf{\gpt}}\\
\midrule
Zero-shot CoT & 62.5 & 37.2 & 78.1 & 78.5 & 76.8 & 66.6 \\
Self-Refine (Best Iter) & 61.1 & 37.4 & 77.9 & 78.4 & 77.1 & 66.4 \\
\rowcolor{blue!7}Best-of-$k$ ($k = 120$) & 70.1 & 50.6 & 87.7 & 90.5 & 87.8 & 77.3 \\
$k$-way SC ($k = 120$) & 70.4 & 51.2 & 86.9 & 89.8 & 87.6 & 77.1 \\
Self-Refine + $k$-way SC (Best Iter) & 70.1 & 49.4 & 88.1 & 88.1 & 84.5 & 76.0  \\\midrule
\method{} (Iter=1) & \textbf{73.7} & 57.2 & 89.4 & 91.1 & 90.1 & 80.3 \\
\method{} (Iter=2) & 73.3 & 57.8 & \textbf{90.1} & 91.1 & \textbf{90.9} & 80.6 \\
\method{} (Iter=3) & 73.6 \plus{3.5} & \textbf{58.6} \plus{8.0} & \textbf{90.1} \plus{2.4} & \textbf{91.4} \plus{0.9} & \textbf{90.9} \plus{3.1} & \textbf{80.9} \plus{3.6} \\
\bottomrule
\end{tabular}}
\caption{Performance comparison of methods and models. \plus{x} is compared to the strongest baseline (Best-of-$k$) shown in \setlength{\fboxsep}{1pt}\colorbox{blue!7}{blue}.
Across models and datasets, \method{} consistently improves. Notably, \method{} surpasses all baselines after the \textit{first iteration} of refinement, even when the baselines use a larger sample size ($k = 120$). 
}
\label{tab:main}
\vspace{-10pt}
\end{table*}

\subsection{Main Results}
\label{ssec:main}
\begin{figure}
  \begin{center}
    \includegraphics[width=0.42\textwidth]{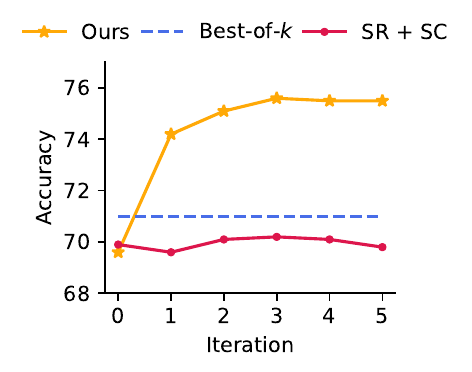}
  \end{center}
   \vspace{-15pt}
    \caption{Comparison with baselines across iterations (avg. of 5 datasets with $k=40$). Full results: \cref{tab:more}.}
    \label{fig:iteration}
\end{figure}

\textbf{\method{} outperforms all baselines at the first iteration.}
We present our main results in \cref{tab:main}. 
First, one iteration of \method{} already outperforms all baselines.
Compared to aggregation-based methods, which generate multiple responses for each problem without refinement, \method{} improves over Best-of-$120$ by $3.2\%$ (absolute) averaged across the five datasets on Llama-3-8B, despite using $2\times$ fewer samples.
Note that our method's first iteration only involves $40$ samples for easy problems and $40$ refined chains for the \emph{subset} of hard problems, making our $k=55$ on average. 
When compared to $120$-way SC, our method shows an even greater average improvement of $3.3\%$ on Llama-3-8B and $3.2\%$ on GPT-3.5.
Turning to refinement-based methods, we run them with up to $5$ iterations and only report the best in \cref{tab:main} (denoted as ``Best Iter''), leaving a more detailed comparison in \cref{fig:iteration} and \cref{tab:more}.
On average, \method{} shows $17.1\%$ and $13.5\%$ improvements over SR for Llama-3-8B and GPT-3.5. As SR alone is a weaker baseline without aggregating multiple samples, we also compare to SR + SC, and find that even with its best iteration, \method{} outperforms SR + SC by $5.4\%$ (Llama-3-8B) and $4.9\%$ (GPT-3.5) on average.
This suggests that adaptively addressing challenging instances with targeted refinement improves overall performance, while reducing compute for easy problems.

\noindent\textbf{\method{} continues to improve with more iterations.}
While \method{} already beats \emph{all} baselines after the very first iteration, 
in \cref{tab:main}, we also observe a clear upward trend in performance as the number of iterations increases. 
We illustrate this further in \cref{fig:iteration}, which presents the accuracy across successive iterations. Our comparison includes Best-of-$k$ and SR + SC with $k=40$, with accuracy averaged across five datasets. 
We find that while SR + SC fluctuates around the same range of $70\%$, \method{} continues to improve and stabilize at the third iteration with $75.6\%$ accuracy (with a noticeable $1.4\%$ improvement compared to the first iteration). 
This highlights the importance of our iterative refinement and the ability to overcome insufficient refinement for hard instances, and indicates that the issue of over-refinement does not reappear in \method{} even after more iterations.

\begin{figure}
  \begin{center}
    \includegraphics[width=0.42\textwidth]{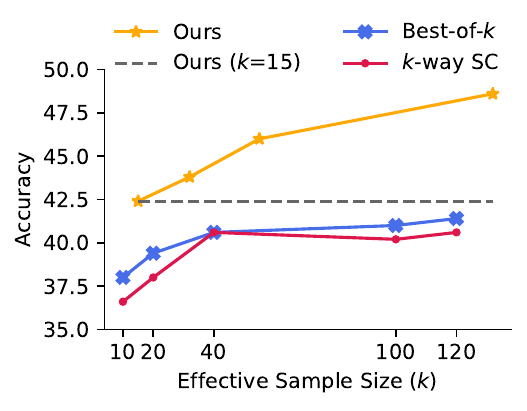}
  \end{center}
    \vspace{-13pt}
    \caption{Comparison of \method{}, $k$-way SC, and Best-of-$k$ with different $k$ on MATH. 
    }
    \label{fig:acc_k}
\end{figure}

\noindent\textbf{\method{} outperforms aggregation-based methods despite using less computation.} 
In \cref{fig:acc_k}, we further compare the cost and performance of \method{} with Best-of-$k$ and SC on MATH and MMLU using Llama-3-8B, studying how performance changes as we increase the number of reasoning chains generated per question $k$. 
Note that we sample $k$ reasoning chains per question for baselines, whereas, in our method, we add $k$ more samples in each iteration for a subset of hard problems, and plot the average number of samples in \cref{fig:acc_k}.
The trend in \cref{fig:acc_k} shows that \method{} consistently outperforms $k$-way SC and Best-of-$k$ at any given $k$. Moreover, while SC saturates and stops improving at around $k=40$, \method{} continues to improve with increasing $k$. 
Notably, \method{} with $k=15$ already performs better than Best-of-$120$ and $120$-way SC, highlighting the cost-effectiveness of our method. We also show that \method{} is also more cost-efficient in terms of token count in \cref{fig:token_count}.

\begin{table*}[ht]
\centering
\begin{tabular}{lcccccc}
\toprule
\textbf{Method} & \textbf{MMLU} & \textbf{MATH} & \textbf{SVAMP} & \textbf{GSM8K} & \textbf{SAT} & \textbf{Avg.} \\
\midrule
Qwen2.5-Math-7B                  & 73.9 & 78.8 & 91.8 & 94.9 & 92.3 & 86.3 \\
$k$-way SC ($k = 40$)            & 81.3 & 87.0 & 95.5 & 97.2 & \textbf{97.3} & 91.7 \\
$k$-way SC ($k = 120$)           & 82.0 & 86.8 & 95.4 & \textbf{97.3} & \textbf{97.3} & 91.8 \\
Best-of-$k$ ($k = 120$)          & 82.6 & 86.0 & 93.4 & 96.9 & 95.2 & 90.8 \\\midrule
\method{}            & \textbf{84.6} & \textbf{91.4} & \textbf{95.8} & \textbf{97.3} & \textbf{97.3} & \textbf{93.3} \\
\bottomrule
\end{tabular}
\caption{\method{} scales with the strength of reward models while also improves stronger base model like Qwen2.5-Math-7B. Here we use Skywork-Reward-Llama-3.1-8B as the ORM and Qwen-Math-PRM-7B as the PRM. Note that all the baselines are using the same models.}
\label{tab:new_models}
\end{table*}

\subsection{Additional Analyses}
\noindent\textbf{\method{} scales with stronger models.}
To evaluate the scalability of \method{} on more recent and capable models, we experiment with Qwen2.5-Math-7B \cite{yang2024qwen25math}, using Skywork-Reward-Llama-3.1-8B \cite{liu2024skywork} as the ORM and Qwen-Math-PRM-7B \cite{yang2024qwen25math} as the PRM. As shown in \cref{tab:new_models}, \method{} achieves the largest gains compared to Self-Consistency and Best-of-$k$ under the same model setup. These results indicate that \method{} not only benefits stronger models like Qwen2.5-Math-7B but also generalizes well across datasets.

\begin{table}[h!]
\vspace{5pt}
\small
\centering
\begin{tabular}{lcc}
\toprule
\bf Method & \textbf{MMLU} & \textbf{MATH} \\
\midrule
Only Address Issue 1 & 64.7 & 44.0 \\
Only Address Issue 2 & 65.9 & 45.4 \\
Only Address Issue 3 & 60.3 & 36.4 \\
MAgICoRE & \textbf{68.9} & \textbf{47.8} \\
\bottomrule
\end{tabular}
\caption{Ablation study on addressing each refinement (c.f. \cref{fig:intro}) issue one at a time.}
\label{tab:issue_ablation}
\end{table}

\noindent\textbf{All three issues must be addressed jointly.} 
To investigate the importance of each refinement issue and component in \method{}, we conduct an ablation study to address each issue individually in \cref{tab:issue_ablation} under the following settings: (1) \textit{Only Address Issue 1 (Excessive Refinement)}: Here, we apply selective refinement only, without PRM step-wise scores for feedback generation and without iterations. (2) \textit{Only Address Issue 2 (Inability to Localize and Address Errors)}: Here, we use PRM scores for feedback generation and refine all instances uniformly (i.e. no selective refinement) for one iteration. (3) \textit{Only Address Issue 3 (Insufficient Refinement)}: Here, we iteratively refine all samples without incorporating PRM scores (i.e. no error localization) and without performing selective refinement. 
The results show that only addressing one single refinement issue at a time leads to a performance drop, highlighting the need for a joint solution as we proposed in \method{}.
We find that only addressing insufficient refinement (Issue 3) causes the highest drop in performance, as it fails to efficiently localize errors (without the help of PRM) and also performs excessive refinement.

\begin{table}[h!]
    \vspace{5pt}
    \small
    \centering
    \begin{tabular}{ccc}
    \toprule
    \textbf{PRM} & \textbf{ORM} & \textbf{Acc.} \\
    \midrule
    MS-7B &  ILM-7B & 47.8 \\
    \midrule
    QM-7B &  ILM-7B & 52.6 \\
    QM-72B &  ILM-7B & 55.4 \\
    \midrule
    MS-7B & SRL-8B & 49.4 \\
    \bottomrule
    \end{tabular}
    \caption{Performance of \method{} with different RMs, which can be swapped in without modification.}
    \label{tab:rms}
\end{table}

\noindent\textbf{Modularity of \method{}.}
In \cref{tab:main} we report performance using InternLM-7B (ILM-7B) as the ORM and Math-Shepherd-7B (MS-7B) as the PRM. 
Here, we illustrate the modularity of \method{} by incorporating different ORM and PRMs; note that this can be done without changes to the code.
In \cref{tab:rms}, we report the performance of \method{} on MATH when using different ORMs and PRMs, holding the other fixed.
We test Qwen-Math PRM 7B and 72B \citep[QM-7B and 72B;][]{zhang2025lessons} as PRMs and Skywork-Reward-Llama-3.1-8B \citep[SRL-8B;][]{liu2024skywork} as an alternate ORM.
In all cases, we find that \method{} benefits from other RM selections, and that these changes can be made trivially.

\begin{table}[h!]
\vspace{5pt}
\small
\centering
\begin{tabular}{lcc}
\toprule
\textbf{Method} & \textbf{MMLU} & \textbf{MATH} \\
\midrule
Aggregation-Only & 64.7 & 44.0 \\ 
Refinement-Only   & 60.9 & 38.8 \\
\method{} & \textbf{67.3} & \textbf{46.0}\\
\bottomrule
\end{tabular}
\caption{Comparison when uniformly adopting aggregation (i.e., Weighted SC) or refinement to \emph{all instances}.}
\label{tab:contrib}
\end{table}

\noindent\textbf{Selective refinement avoids over-correcting and improves overall performance.} 
\label{sec:refine}
In \cref{sec:intro}, we noted that excessive refinement could potentially hurt performance by flipping correct answers to incorrect ones. Here, we provide a quantitative analysis of this claim. 
Recall that we have two methods: coarse aggregation (Weighted SC) and fine refinement (multi-agent iteration) which we apply \textit{selectively} depending on predicted problem difficulty (c.f. \cref{sec:stage1}).
In \cref{tab:contrib}, we measure the performance of each method when applied uniformly to \emph{all} instances, regardless of the problem difficulty.
We find that uniformly applying refinement actually degrades performance; comparing Weighted SC (the ``Aggregation-Only'' in row 1) to refinement-only (row 2), we see that refining all samples leads to $3.8\%$ and $5.2\%$ drops on MMLU and MATH, respectively, pointing to the over-correction issue.
Conversely, one iteration of our selective refinement (row 3) targets only the challenging instances where the weighted majority vote is unlikely to succeed, resulting in up to $2.6\%$ improvement compared to uniformly applying aggregation (row 1). This demonstrates that our selective refinement not only avoids over-correction but also enhances overall performance by effectively allocating more resources to harder problems.

\begin{table}[h!]
    \vspace{5pt}
    \small
    \centering
    \setlength{\tabcolsep}{2.5pt}
    \begin{tabular}{lcc}
    \toprule
    \textbf{Refinement Variants} & \textbf{MMLU} & \textbf{MATH}\\ 
    \midrule
    LLM Self-Refinement & 65.9 & 44.4 \\
    Random Step Score & 66.4 & 43.8 \\  
    ORM Score (No Step Score) & 66.8 & 45.2 \\
    Ours (PRM Step Score) & \textbf{67.3} & \textbf{46.0} \\ 
    \bottomrule
    \end{tabular}
    \caption{Refinement variants in \method{}. Using PRM scores for refinement performs the best.
    }
    \label{tab:llmrefine}
\end{table}
\noindent\textbf{PRM-based feedback enables better refinement.}
Having demonstrated that selectively applying refinement is crucial for achieving improvements, we now compare the refinement process with and without using a PRM. To this end, \emph{without using a PRM}, we ask the LLM to generate an updated solution based on its own previous reasoning, referring to this as LLM Self-Refinement. 
Compared to \method{} in row 4 of \cref{tab:llmrefine}, using LLM's self-refinement (row 1) results in an average drop of $1.5\%$, indicating that using the LLM for refinement is less effective than using a PRM. 
To further examine how sensitive the refinement process is to the score quality, in row 2, we replace the actual PRM scores with random scores. The result is worse than row 4, indicating that PRM scores help in localizing errors.
Finally, we test whether the global ORM score can offer a similar advantage as using the local PRM score. Result in ow 3 shows that it performs slightly worse than using the PRM score, suggesting that while global correctness is also a strong signal, local correctness scores help identify and correct errors more effectively. 
\label{ssec:llmrefine}

\begin{table}
\centering
\small
\begin{tabular}{lcc}
\toprule
\bf Method & \bf ARC & \bf Date \\
\midrule
Zero-shot & 66.5 & 52.5 \\
40-way SC & 85.5 & 72.5 \\
120-way SC & 86.0 & 72.5 \\
MAgICoRE (Iter = 1) & 87.5 & 79.5 \\
MAgICoRE (Iter = 2) & 88.0 & 79.5 \\
MAgICoRE (Iter = 3) & \textbf{88.5} & \textbf{80.5} \\
\bottomrule
\end{tabular}
\caption{\method{} also generalizes to commonsense reasoning and logical reasoning tasks.}
\label{tab:generalization}
\end{table}

\noindent\textbf{\method{} generalizes to other domains.}
\label{sec:generalization}
\cref{tab:main} shows the benefits of \method{} on math reasoning; however, LLMs have been applied to a wide variety of tasks beyond math. 
Here, we explore expanding \method{} to other domains, specifically to a commonsense reasoning task: ARC-Challenge \citep{clark2018think}, and a logical reasoning task: Date Understanding \citep{srivastava2022beyond}. 
We sample 200 instances from each dataset and use GPT4o-mini as a PRM for the experiments, as existing standalone PRMs generally only exist for math. 
Specifically, we prompt GPT4o-mini to provide step-wise correctness scores \emph{without any textual explanations} or reasoning, acting the same as a PRM. The prompt is provided in \cref{sec:gpt4o_prompt}.
This approach ensures that our agents do not have access to explanations from a stronger model.
We conduct this experiment with \llama~as the base LLM. \cref{tab:generalization} shows that \method{} transfers to commonsense and logical reasoning, outperforming $120$-say SC by $2.5$\% and $8.0$\%, respectively. 

\begin{table}[h!]
\vspace{5pt}
\centering
\small
\begin{tabular}{lc}
\toprule
\bf Method & \bf Accuracy \\
\midrule
Zero-shot & 72.0 \\
40-way SC & 79.2 \\
40-way SC + PRM & 79.4 \\
MAgICoRE (Iter = 1) & 80.2 \\
MAgICoRE (Iter = 2) & \textbf{80.4} \\
MAgICoRE (Iter = 3) & \textbf{80.4} \\
\bottomrule
\end{tabular}
\caption{\method{} can also improve GPT4o-mini.}
\label{tab:gpt4o}
\vspace{-5pt}
\end{table}

\paragraph{\method{} also improves stronger models like GPT4o-mini.}

\cref{tab:main} shows results with GPT-3.5-Turbo; here, we show that \method{} scales to its stronger variant as well. 
Specifically, we run \method{} using GPT4o-mini on a subset of MATH data. 
Due to the high cost and the fact that \cref{fig:acc_k} shows \emph{decreasing} performance at $k=120$ for MATH, we only compare to the 40-way SC with the weighted variation that incorporates PRM scores for vote weighting (40-way SC + PRM). 
\cref{tab:gpt4o} demonstrates that \method{} can also enhance stronger model's performance, albeit with a smaller margin of improvement compared to Llama3-8B and GPT-3.5 shown in \cref{tab:main}.

\section{Conclusion}
Building on the observation that different problems require varying amounts of computation, we introduced \method{}, a method that adaptively allocates more computational resources to more challenging problems and selectively applies refinement where appropriate, i.e., on harder problems.
\method{} addresses three key issues in refinement: excessive refinement on easy examples, the inability of LLMs to detect and correct errors, and insufficient refinement on hard instances. Our approach tackles these issues by employing both global and local reward models to decide which samples to refine. We then incorporate local correctness scores to generate targeted feedback and an iterative multi-agent communication framework to refine solutions for hard problems.
Results across five math datasets and two models show that our coarse-to-fine method consistently outperforms both coarse-grained aggregation and fine-grained refinement alone at any given budget, and even outperforms baselines using substantially more computation.
In our ablations, we demonstrate the importance of selective refinement, showing that performance generally drops when refining all samples uniformly. 
We also highlight the role of iteration in our framework, showing increased performance across iterations even as baselines stagnate. 

\section*{Limitations}
Like all test-time scaling, \method{} improves performance by adding computation via additional samples, trading some efficiency for better performance. 
We show that \method{} makes better use of additional compute than the baselines by performing targeted refinement and thus better using inference-time compute; indeed, while baselines like Best-of-$k$ and Self-Consistency stagnate with additional compute, \method{} continues to improve. 
Nevertheless, our method increases the computational cost of inference, and relies on starting with a base number of samples to establish the difficulty and quality of existing solutions. 
In addition to requiring multiple solutions, \method{} uses feedback from both ORMs and PRMs to improve refinement. 
These models must be separately trained to provide rewards for a given domain and therefore do not exist for all problem types. 
However, we also note that \method{} is modular, and thus allows for newer and better ORMs and PRMs to be swapped in as they become available. 
\method{} is designed to improve the reasoning of LLMs, and thus has no additional risks beyond those inherent to LLMs generally. 

\section*{Acknowledgments}
We would like to thank the anonymous reviewers for their feedback. 
This work was supported by NSF-CAREER Award 1846185, NSF-AI Engage Institute DRL-2112635, DARPA MCS Grant N66001-19-2-4031, a Capital One Research Award, a Cisco Research Award, and a Google PhD Fellowship, and the Accelerate Foundation Models Research program. The views contained in this article are those of the authors and not of the funding agency.

\bibliography{custom}

\appendix

\input{appendix}

\end{document}

%% file: appendix.tex
\section*{Appendix}
\section{Related Work}
\textbf{Improving Reasoning by Aggregation.} 
Self-Consistency \citep[SC;][]{wang2022self} generates $k$ reasoning chains and marginalizes over the chains to obtain answer clusters; the most frequent answer is selected as the final prediction.
While simple and effective, it generates $k$ solutions \emph{for every sample}, as both past work and our work show that SC saturates when $k$ increases \citep{chen2024llmcallsneedscaling,li2024agentsneed}. 
Several studies adaptively determine the number of samples ($k$) required \emph{for each instance} to address this \citep{aggarwal2023let,li2024escape}. 
However, the performance of such approaches is upper-bounded by SC -- they address the cost issue but do not enhance overall performance. 
To surpass SC, \citet{yin2024aggregation} propose using the LLM to evaluate answer clusters, taking into account both frequency and the LLM-evaluated quality of the answers.
Instead, we propose using external RMs to decide between coarse-grained aggregation and fine-grained refinement, identify errors, and aid refinement; this allows us to improve over aggregation or refinement alone.

\noindent\textbf{LLM-based Verification and Refinement.} 
Past work mostly uses RMs for verification purposes \citep{li2023making, khalifa2023grace, cobbe2021training, lightman2023let}. 
\citet{havrilla2024glore} considers local correctness for refinement in a non-adaptive way and requires specific data curation with fine-tuning, while our work is adaptive and uses off-the-shelf global and local models.
Another line of work has proposed using the LLM itself as a verifier, in place of an RM \citep{liu2024large, zhang2024accessing, aggarwal2023let, madaan2023self}.
However, recent work shows the inability of LLMs to ``self-verify'' their own reasoning~\citep{huang2024large, stechly2023gpt4, kamoi2024can,tyen2024correct,kamoi2024can}. 
Hence, \method{} uses external global and local reward models \citep{wang2023math, cai2024internlm2} for selective coarse-to-fine refinement.
\citet{shridhar2024art} trained specific models to decide when to refine and when to trust refined solutions. 
This contrasts with our method, where the decision to refine is based on a coarse-to-fine resource allocation method that differentiates easy from hard problems (for which we use global and local reward models), and where refinement is done based on off-the-shelf models. 
Past work has also used RMs to guide MCTS search for math problems \citep{guan2025rstarmathsmallllmsmaster}.
We do not compare to such methods, as their use of multiple rollouts makes generation-matched comparisons like the kind we do challenging.
Moreover, while \method{} explores how reward models can effectively address issues in refinement, MCTS-based methods primarily investigate how reward models can guide the search toward the final answer. These distinct goals make direct comparison less meaningful.

We also iteratively refine guided by global correctness to ensure sufficient refinement, whereas \citet{shridhar2024art} refine only once.
\citet{wadhwa2024dcr} propose a multi-agent detect-critique-refine pipeline for generation tasks. 
\method{} instead focuses on reasoning tasks and uses external RMs for selective coarse-to-fine refinement (whereas \citet{wadhwa2024dcr}'s detection uses the same metric as their evaluation, which is infeasible in reasoning where the metric -- accuracy -- requires access to the gold answer).

\noindent\textbf{Multi-Agent Systems with LLMs.}  
LLMs can be used in multi-agent systems, where the agents interact, collaborate, and compete \citep{wang-etal-2024-unleashing, lu2024llm, feng-etal-2024-dont}. 
Related to our work, one line of multi-agent research focuses on structured debates or discussions between LLM agents, where the interaction helps refine and improve previously generated solutions \citep{du2023debate, liang2023encouraging, yin2023exchange, chen2024reconcile}. 
These studies show improvements over single-agent systems, but a major challenge in multi-agent systems is achieving a correct consensus among LLMs; external feedback can help prevent this consensus from aligning with the agents' internal and possibly erroneous outputs. 
Therefore, \method{}'s multi-agent refinement incorporates external RMs for more objective scoring, enabling the generation of targeted feedback for better refinement.

\section{Self-Consistency and Weighted Self-Consistency}
\label{ssec:wsc}
Self-Consistency \citep{wang2022self} is a popular decoding method that uses majority voting to aggregate predictions from different reasoning chains, thus marginalizing over chains. 
It generates $k$ solutions per question and selects the most frequent final answer from these samples. 
While simple and effective, this method assigns uniform weight to each reasoning chain, which fails to account for the quality of each solution. 
To address this limitation, \citet{li2023making} propose Weighted Self-Consistency, accounting for each solution's quality. 
Formally, both Self-Consistency and Weighted Self-Consistency choose a final answer via: 
    \[ \hat{y} = \arg\max_{y} \sum_{i=1}^{k} \mathbbm{1}_{y_i=y} \mathcal{V}(q; r_i) \]
where $\mathcal{V}(\cdot)$ is a constant $1$ in Self-Consistency and the quality measurement (e.g., RM score) in Weighted Self-Consistency. 
In \method{}'s final answer selection, we use the sum of the solution-level scores generated by both ORM and PRM as $\mathcal{V}(\cdot)$. Throughout \method{}, in cases where we need solution-level PRM scores (compatible with ORM scores), we accumulate the PRM step scores by taking their product \citep{sun2024easytohardgeneralizationscalablealignment}, so that the aggregated PRM score corresponds to a solution. 

\section{Details of the Conditions}
\label{sec:cond}
\paragraph{Condition 1: Is the Majority Answer of High Quality?}
Given a problem $q$, to determine the difficulty of the problem at hand, the Solver generates $k$ solutions $R = \{r_1, \ldots, r_k\}$ and final answers $A = \{a_1, \ldots, a_k\}$ per question and cluster the solutions by their final answer.
This produces a partition $\mathcal{A}$ with elements $\mathcal{A}_i$, where $\mathcal{A}_i = \{r_j \in R\ |\ a_j = a_i\}$.
The majority cluster $\mathcal{A}_g$ has the most ``votes'', i.e., 
$\mathcal{A}_g = \mathrm{argmax}_{i\in |\mathcal{A}|} |\mathcal{A}_i|$. 
We evaluate the majority answer quality by both ORM and PRM separately but with the same procedure, as described below.
First, we score every reasoning chain $r_i$ within the majority cluster $\mathcal{A}_g$. Both ORM and PRM are able to produce a solution-level score, which we denoted as $\mathcal{S}^{\textsc{rm}}_{i}$. Note that we perform this check using ORM and PRM separately, but for simplicity, we use the same notation for solution-level score, which either comes from the ORM or the PRM.
We calculate the average of the solution-level scores from the majority group:
\[
\mathcal{S}^{\textsc{rm}}_{avg} = \frac{1}{|\mathcal{A}_g|} \sum_{i=1}^{|\mathcal{A}_g|} \mathcal{S}^{\textsc{rm}}_{i}
\]
This average score informs us of the majority answer's quality.
To set a threshold, we normalize $\mathcal{S}^{\textsc{rm}}_{avg}$ by using the sample average RM scores (by computing $\mathcal{S}^{\textsc{rm}}_{i}$ for each sample and then take the average of these scores). Importantly, this process does not require any labeled data.
After normalization, if the average reward of the majority group $\mathcal{S}^{\textsc{RM}}_{avg} \geq 0$, indicating that the quality of the majority answer is high, Condition 1 will be \emph{true}. 
Otherwise, if $\mathcal{S}^{\textsc{RM}}_{avg} < 0$, Condition 1 will be \emph{false}, suggesting that even the most frequent answer is of poor quality and that the instance might benefit from refinement. 

\paragraph{Condition 2: Are Reward Models' Answer Confidence High?}
Besides the quality of the majority answer, we also consider whether the RMs are confident enough in any single answer among the answer clusters. Again we evaluate both ORM and PRM's answer confidence separately but with the same procedure, as described below.
First, the answer distribution is formed by (1) the frequency of each unique answer and (2) the total RM score of each answer cluster. We estimate the RM's confidence according to this distribution. If the distribution is concentrated, meaning that only one answer cluster stands out, the RM's answer confidence is treated as high. 
Conversely, if the distribution is diffused and the clusters' scores are more uniform, then there is no single answer for which the RM has high confidence, i.e., the RM's confidence is low. 
This motivates a targeted step-wise refinement process to select a more definite answer.
Again we use both ORM and PRM to generate the solution-level score $\mathcal{S}^{\textsc{RM}}_{i}$. Given the $k$ reasoning chains generated along with the solution-level score, 
we compute the RM's answer confidence (denoted as $C$) using the entropy of the answer cluster weighed by the RM scores, passing the result through a sigmoid function to normalize it onto $[0,1]$. 
Formally, the calculation of the entropy can be expressed as: 
\[
H = - \sum_{i=1}^n p(\mathcal{A}_i) \log p(\mathcal{A}_i),
\]
\vspace{-15pt}
\[
p(\mathcal{A}_i) = \frac{\sum_{i = 1}^{|\mathcal{A}_i|} \mathcal{S}^{\textsc{RM}}_{i}}{\sum\limits_{\mathcal{A}_j \in \mathcal{A}} \sum_{k = 1}^{|\mathcal{A}_j|} \mathcal{S}^{\textsc{RM}}_{k}}
\]
where $n$ is the number of unique answers among the $k$ chains, $\mathcal{A}_i$ is the $i$-th answer cluster (a set of reasoning chains leading to the same answer) and $\mathcal{A}$ is the set of all clusters. 
Each answer in a cluster is weighed by its unnormalized solution-level score $\mathcal{S}^{\textsc{RM}}_{i}$. 
To normalize entropy onto a confidence scale, we invert it so that high entropy corresponds to low confidence. We then apply a sigmoid function $\sigma (.)$, mapping the values to the range $[0,1]$: $C = \sigma(\alpha * (1-H))$. We set $\alpha$ to $2$ to let the distribution stretch more evenly between 0 and 1. This transformation establishes $0.5$ as a natural threshold for differentiating low and high confidence, thereby eliminating the need for any threshold tuning.
That is, if an instance has $C \geq 0.5$, Condition 2 is \emph{true}, meaning that the RMs are confident on a single answer cluster.
Otherwise, if $C < 0.5$, Condition 2 is \emph{false}, suggesting that the RMs' uncertainty among the $k$ chains is high, necessitating a finer refinement.

\section{Additional Experimental Results}

\begin{table*}[t!]
\centering
\resizebox{\textwidth}{!}{%
\begin{tabular}{lcccccc}
\toprule
\textbf{Method} & \textbf{MMLU} & \textbf{MATH} & \textbf{SVAMP} & \textbf{GSM8K} & \textbf{SAT} & \textbf{Avg.} \\ \midrule
120-way SC & 63.0 & 40.6 & 89.8 & 90.3 & 70.5 & 70.8 \\ 
120-way SC + PRM \cite{li2023making} & 65.4 & 44.6 & 90.8 & 90.7 & 72.5 & 72.8 \\
Self-correct + 120-way SC \citep{kim2024language} & 62.1 & 38.6 & 86.2 & 88.1 & 65.6 & 68.1 \\
Least-to-Most + 120-way SC \citep{zhou2022least} & 62.6 & 40.6 & 89.0 & 90.3 & 68.9 & 70.3 \\
Multi-Agent Debate + SC \citep{du2023debate} & 64.6 & 41.0 & 89.6 & 90.8 & 72.5 & 71.7 \\ \midrule
MAgICoRE (Iter=1) & 67.3 & 46.0 & \textbf{91.4} & 91.1 & 75.0 & 74.2 \\
MAgICoRE (Iter=2) & 68.4 & 47.2 & 91.1 & \textbf{92.3} & 76.4 & 75.1 \\
MAgICoRE (Iter=3) & \textbf{68.9} & \textbf{47.8} & 91.3 & 91.6 & \textbf{78.2} & \textbf{75.6} \\ \bottomrule
\end{tabular}}
\caption{Performance comparison with additional baselines using \llama. Notably, \method{} with only one iteration outperforms all baselines despite using fewer samples.}
\label{tab:additional_baselines}
\end{table*}

\paragraph{Comparison with additional baselines.}
In addition to \cref{tab:main}, we also compare with the following baselines: 
(1) 120-way SC + PRM: The product of step-wise PRM scores is used as the solution-level score. This score is then employed for weighted Self-Consistency, following \cite{li2023making}.  
(2) Self-correct + 120-way SC: We use the ``Self-Correct RCI'' prompt from \cite{kim2024language} to generate 120 solutions per question, which are subsequently aggregated using Self-Consistency.  
(3) Least-to-Most + 120-way SC: We use the zero-shot Least-to-Most prompt from \cite{zhou2022least} to generate 120 solutions per question, followed by aggregation via Self-Consistency.  
(4) Multi-Agent Debate + SC: Following \citet{du2023debate}, we conduct a three-agent debate over four rounds, repeating this process ten times. The final answers from these ten debates are aggregated using Self-Consistency, yielding 120 generations per question.

We use~\llama~as the base model. Results show that a single iteration of \method{} already outperforms methods that rely on PRM for aggregation (120-way SC + PRM), as well as approaches like Self-Correction, advanced prompting, and multi-agent debate. 
On average, \method{} outperforms 120-way SC + PRM by $2.8\%$ despite using fewer samples, highlighting the limitations of using PRM solely for aggregation. Additionally, \method{} exceeds Least-to-Most by $5.3\%$, showcasing superior adaptability to problem difficulty. Finally, \method{} surpasses Multi-agent Debate by $3.9\%$, indicating that our aggregation and refinement mechanisms scale more effectively at test time.

\begin{table}
\centering
\resizebox{0.35\textwidth}{!}{%
\begin{tabular}{lcc}
\toprule
\textbf{Aggregation} & \textbf{MMLU} & \textbf{MATH} \\
\midrule
ORM-Only   & 66.9 & 45.4 \\
PRM-Only & 66.1 & 45.0 \\ 
Both & \textbf{67.3} & \textbf{46.0}\\
\bottomrule
\end{tabular}}
\label{tab:ormprm_ablation}
\caption{Ablation study on the final answer selection, using ORM-only, PRM-only or both.}
\end{table}

\begin{table}
    \centering
    \small
    \begin{tabular}{lcc}
    \toprule
     & \textbf{MMLU} & \textbf{MATH}\\ 
    \midrule
    Joint Roles & 66.7 & 44.8 \\ 
    Distinct Agents (Ours) & \textbf{67.3} & \textbf{46.0} \\ 
    \bottomrule
    \end{tabular}
    \label{tab:multiagent}
    \caption{\method{}'s separation of the Reviewer and Refiner roles is more effective than combining them into a single role.}
\end{table}

\paragraph{Separating Reviewer and Refiner roles outperforms combining these roles.}
In \cref{tab:multiagent}, we examine the effects of combining the roles of Reviewer and Refiner by merging their prompts, instructing the model to simultaneously generate both feedback and a refined solution. This method is referred to as ``Joint Roles''. 
In \method{}, the Reviewer and Refiner have distinct, clearly defined roles,
which we refer to as the ``Distinct Agents'' approach. As before, the performance comparison is based on the first iteration, with all other variables held constant. Our findings show that maintaining separate roles (as in our multi-agent setup) leads to better performance, with the ``Joint Roles'' configuration resulting in a $0.6\%$ drop in MMLU and a $1.2\%$ decrease in MATH. The larger drop in MATH suggests that its problems are more complex and often require extended reasoning, making the combined roles less effective, whereas maintaining separate roles proves to be more beneficial.

\paragraph{Ablations on reward models for final answer selection.}
We report \method{} up to three iterations in \cref{tab:main} and only report the best-performing iteration of Self-Refine + $k$-way SC. Here, we provide extended results in table \cref{tab:more}.
We also conducted another ablation study to evaluate the performance when using ORM, PRM, or a the summation of both scores for final answer selection. As shown in \cref{tab:ormprm_ablation}, utilizing ORM's global correctness score yields better results than aggregating PRM's local correctness score. However, the best performance is achieved when both scores are combined for the final answer aggregation.

\begin{table}
\centering
\resizebox{0.4\textwidth}{!}{%
\begin{tabular}{lc}
\toprule
\textbf{Refinement Variants}                & \textbf{Accuracy} \\ \midrule
No feedback (LLM self-refine)   & 48.30             \\ 
Random PRM score                & 49.60             \\ 
PRM predicted score             & 51.20             \\ 
Oracle PRM score                & \textbf{52.40}             \\ \bottomrule
\end{tabular}}
\label{tab:prm_oracle}
\caption{Comparison of different refinement variants in \method{}.}
\end{table}

\paragraph{Reliable step-wise scores enable LLM refinement.}
To compare with an oracle PRM, we sample 500 instances from the Math-Shepherd dataset \citep{wang2023math}, which includes gold label correctness for each step. Besides the three settings we evaluated in \cref{tab:llmrefine}, we also evaluate the oracle PRM score, where feedback uses the gold correctness labels. \cref{tab:prm_oracle} shows that the oracle PRM score performs the best, followed by the predicted PRM score, suggesting that given reliable stepwise scores, LLMs can effectively refine their solutions and improve.

\begin{table}
    \small
    \centering
    \small
    \setlength{\tabcolsep}{2.5pt}
    \begin{tabular}{lcc}
    \toprule
    \textbf{Criterion for Refinement} & \textbf{MMLU} & \textbf{MATH}\\ 
    \midrule
    Prompt (classification) & 65.2 & 45.0 \\ 
    Prompt (confidence) & 64.7 & 44.4 \\
    Condition 1 only & 66.4 & 43.6 \\ 
    Condition 2 only & 66.1 & 44.2 \\ 
    Cond. 1 \& Cond. 2 & \textbf{67.3} & \textbf{46.0}\\ 
    \bottomrule
    \end{tabular}
    \caption{Different ways of detecting hard problems (i.e. criterion for refinement). Our two conditions, when used together, are the most effective.} 
    \label{tab:verification}
\end{table}

\begin{table}[h!]
    \small
    \centering
        \setlength{\tabcolsep}{2.5pt}
    \resizebox{0.44\textwidth}{!}{%
    \begin{tabular}{lccc}\toprule
       & P & R & F1 \\\midrule
       Random  & 68.4 & 49.6 & 57.5 \\
       Prompt-based (classification) & 65.9 & 10.3 & 17.8 \\
       Prompt-based (confidence) & 0.0 & 0.0 & 0.0 \\
       \method{} & \textbf{86.3} & \textbf{67.6} & \textbf{75.8} \\
       \bottomrule
    \end{tabular}}
    \caption{The Precision (P), Recall (R) and F1 of the model predicted problem difficulty.}
    \label{tab:difficulty}
\end{table}

\paragraph{Effectiveness of the two conditions for classifying problem difficulty.}
In \method{}, we use reward models to classify each instance as easy or hard. Given that the RMs are also fine-tuned LLMs, we investigate whether prompting the LLM to perform this classification directly could replace the external RMs.
We compare two settings in the first two rows, where we prompt \llama~to evaluate the difficulty of an instance. In the first setting (classification), the LLM generates a binary label. In the second setting (confidence), it produces a confidence score ranging from $0$ to $1$, indicating whether refinement is required -- that is, whether the example is easy or hard.
Results in \cref{tab:verification} show that the LLM is less effective at determining instance difficulty compared to a reward model, as evidenced by a performance drop of $1.6\% - 2.6\%$.
In rows 3 and 4, we also examine the performance when only one of the conditions of \method{} (c.f. \cref{sec:stage1}) is used to decide difficulty. Specifically, when only condition 1 is applied, an instance is classified as hard if the majority answer's quality is low. Conversely, when only condition 2 is applied, an instance is classified as hard if the RM's answer confidence is low, regardless of the majority answer's quality. Results indicate that while each condition individually outperforms LLM self-verification, combining both yields the best performance.
Indeed, in \cref{sec:difficulty}, we find that \method{}'s assessment of problem difficulty shows the highest agreement with human-annotated labels.

\begin{table*}[]
\centering
\begin{tabular}{lcccccc} 
\toprule
& \textbf{MMLU} & \textbf{MATH} & \textbf{SVAMP} & \textbf{GSM8K} & \textbf{SAT} & \textbf{Avg.} \\
\midrule
\multicolumn{7}{c}{\textbf{\llama}}\\
\midrule
Zero-shot CoT & 50.4 & 24.2 & 72.4 & 80.1 & 58.2 & 57.1 \\\midrule
Self-Refine (Iter=1) & 49.6 & 24.6 & 72.0 & 79.0 & 57.7 & 56.3 \\
Self-Refine (Iter=2) & 50.2 & 23.8 & 72.8 & 79.6 & 59.3 & 57.1 \\
Self-Refine (Iter=3) & 49.8 & 24.0 & 72.6 & 79.6 & 59.6 & 57.1 \\\midrule
\rowcolor{blue!7}Best-of-$k$ ($k = 120$) &  62.6 & 41.4 & 88.7 & 90.1 & 72.4 & 71.0 \\
$k$-way SC ($k = 120$) & 63.0 & 40.6 & 89.8 & 90.3 & 70.5 & 70.8 \\ \midrule
Self-Refine + $k$-way SC (Iter=0) & 62.1 & 40.4 & 88.6 & 90.1 & 68.2 & 69.9 \\
Self-Refine + $k$-way SC (Iter=1) & 61.3 & 40.6 & 88.9 & 89.7 & 67.7 & 69.6 \\
Self-Refine + $k$-way SC (Iter=2) & 62.7 & 40.0 & 88.9 & 90.1 & 68.6 & 70.1 \\
Self-Refine + $k$-way SC (Iter=3) & 62.3 & 41.0 & 89.2 & 90.3 & 68.0 & 70.2 \\
Self-Refine + $k$-way SC (Iter=4) & 62.1 & 41.4 & 89.2 & 90.1 & 67.7 & 70.1 \\
Self-Refine + $k$-way SC (Iter=5) & 62.7 & 40.4 & 88.6 & 89.7 & 67.7 & 69.8 \\
\midrule
\method{} (Iter=1) & 67.3 & 46.0 & \textbf{91.4} & 91.1 & 75.0 & 74.2 \\
\method{} (Iter=2) & 68.4 & 47.2 & 91.1 & \textbf{92.3} & 76.4 & 75.1 \\
\method{} (Iter=3) & \textbf{68.9} & 47.8 & 91.3 & 91.6 & \textbf{78.2} & \textbf{75.6} \\
\method{} (Iter=4) & \textbf{68.9} & \textbf{48.0} & 91.3 & 91.1 & \textbf{78.2} & 75.5 \\
\method{} (Iter=5) & 68.4 & \textbf{48.0} & 91.1 & 91.6 & \textbf{78.2} & 75.5 \\
\midrule
\multicolumn{7}{c}{\textbf{\gpt}}\\
\midrule
Zero-shot CoT & 62.5 & 37.2 & 78.1 & 78.5 & 76.8 & 66.6 \\\midrule
Self-Refine (Iter=1) & 62.4 & 37.4 & 77.7 & 77.4 & 77.3 & 66.4 \\ 
Self-Refine (Iter=2) & 61.6 & 37.6 & 78.6 & 77.9 & 76.9 & 66.5 \\
Self-Refine (Iter=3) & 61.1 & 37.4 & 77.9 & 78.4 & 77.1 & 66.4 \\\midrule
\rowcolor{blue!7}Best-of-$k$ ($k = 120$) & 70.1 & 50.6 & 87.7 & 90.5 & 87.8 & 77.3 \\
$k$-way SC ($k = 120$) & 70.4 & 51.2 & 86.9 & 89.8 & 87.6 & 77.1 \\\midrule
Self-Refine + $k$-way SC (Iter=0) & 69.4 & 49.8 & 86.9 & 88.1 & 85.6 & 76.0 \\
Self-Refine + $k$-way SC (Iter=1) & 69.8 & 49.0 & 87.1 & 88.3 & 85.0 & 75.8 \\
Self-Refine + $k$-way SC (Iter=2) & 70.1 & 49.4 & 88.1 & 88.1 & 84.5 & 76.0 \\
Self-Refine + $k$-way SC (Iter=3) & 69.6 & 48.8 & 87.3 & 87.8 & 85.2 & 75.7 \\
Self-Refine + $k$-way SC (Iter=4) & 69.8 & 48.4 & 87.1 & 87.1 & 85.0 & 75.5 \\
Self-Refine + $k$-way SC (Iter=5) & 69.6 & 48.6 & 87.3 & 87.4 & 84.5 & 75.5 \\
\midrule
\method{} (Iter=1) & \textbf{73.7} & 57.2 & 89.4 & 91.1 & 90.1 & 80.3 \\
\method{} (Iter=2) & 73.3 & 57.8 & \textbf{90.1} & 91.1 & \textbf{90.9} & 80.6 \\
\method{} (Iter=3) & 73.6 & \textbf{58.6} & \textbf{90.1} & \textbf{91.4} & \textbf{90.9} & \textbf{80.9} \\
\method{} (Iter=4) & 73.6 & 58.0 & 89.9 & \textbf{91.4} & \textbf{90.9} & 80.8 \\
\method{} (Iter=5) & 73.4 & 57.6 & 89.4 & 91.1 & \textbf{90.9} & 80.5 \\
\bottomrule
\end{tabular}
\caption{Extended version of \cref{tab:main}. 
Here we show all more iterations for Self-Refine + $k$-way SC and \method{}. 
While SR + SC does not show a clear improvement with more iterations, \method{} continues to improve, peaking at the third iteration.}
\label{tab:more}
\end{table*}

\paragraph{Model-Predicted vs. Human-Annotated Problem Difficulty.} 
\label{sec:difficulty}
We analyze the model's prediction of problem difficulty. Specifically, we utilize the MATH dataset, which includes human-annotated difficulty levels ranging from 1 to 5, with higher levels indicating increased problem complexity. 
For our analysis, we split the problems as follows:
(1) Easy: Levels 1 and 2 and (2) Hard: Levels 4 and 5.
We exclude Level 3 problems to create a clearer distinction between easy and hard categories. 
We compare the overlap between our model's predictions and these human-annotated levels. 
We treat hard as the positive label. 
The results are presented in \cref{tab:difficulty}. 
To provide a comparative analysis, we include: (1) a random baseline that assigns easy and hard labels at random, (2) a prompt-based baseline that directly prompts the LLM to classify the problem difficulty, and (3) another prompt-based baseline that prompts the LLM to generate a confidence score when answering, where a confidence score of $\geq 0.5$ is classified as ``easy''.
Results show that our conditions substantially outperform all baselines. Interestingly, the prompt-based methods perform worse than the random baseline, particularly the one relying on confidence scores, which classifies \emph{all problems as easy}; this method scores 0 for both precision and recall since we treat ``hard'' as the positive label, so it has 0 true positives. This suggests that our framework is highly effective at distinguishing true problem difficulty based on the conditions outlined in our methodology.

\begin{figure*}[h!]
    \centering
    \includegraphics[width=\linewidth]{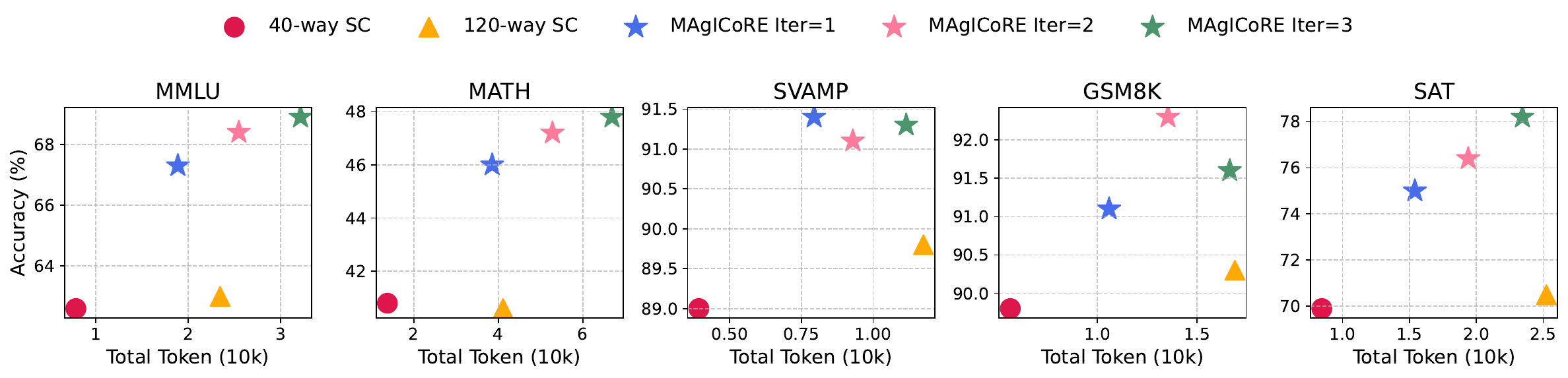}
    \caption{Token count comparison with Self-Consistency across different datasets. Scaling Self-Consistency from $k=40$ to $k=120$ introduces substantial token overhead while providing marginal improvements. In contrast, \method{} demonstrates superior scalability, delivering much higher performance gains with an increased token count. Notably, the first iteration of \method{} consistently outperforms 120-way SC while using fewer tokens.}
    \label{fig:token_count}
\end{figure*}

\paragraph{Token Count Analysis.}
In \cref{fig:acc_k}, we are mainly comparing the number of generations ($k$) per question with the baselines. To provide a more granular analysis, we break down the generations at the token level and compare costs in terms of token counts. The results are detailed in \cref{fig:token_count}. For Self-Consistency, the input tokens are counted only once per question, as it uses the same input to generate $k$ responses. In contrast, the input token count for \method{} includes all prompts across all agents -- Solver, Reviewer, and Refiner. We also include the token count for the ORM and PRM in \method{}. Since the cost of input tokens is typically $0.25\times$ that of output tokens\footnote{See \url{https://openai.com/api/pricing}, \url{https://www.anthropic.com/pricing\#anthropic-api}, and \url{https://ai.google.dev/pricing\#1_5pro}}, we present the normalized total token cost as $0.25 \times$ input $+$ $1 \times$ output.
Results in \cref{fig:token_count} show that (1) scaling Self-Consistency from $k=40$ to $k=120$ largely increases token overhead while yielding marginal improvements. (2) \method{} exhibits superior scalability, achieving substantially higher performance gains with increased token usage. On MMLU, MATH and SAT, we observe a clear upward trend with an increased token count; \method{} consistently improves with additional tokens (unlike SC which tends to stagnate). (3) The first iteration of \method{} outperforms 120-way SC fewer tokens.

\paragraph{Discussion of external reward models.}
External reward models play an important role \method{} and are used in the solutions to all three problems (excessive refinement, inability to localize and address errors, and insufficient refinement). While \method{} does utilize external reward models, our framework is modular and can readily incorporate new reward models as they emerge. 
As the community is actively advancing the performance of reward models evidenced by a benchmark for reward models \citep{lambert2024rewardbench}, \method{} is thus complementary to and enhanced by progress in reward modeling, rather than constrained by it.
While it is possible to train a custom error-identification model, this approach is often data-dependent and prone to obsolescence. In contrast, \method{}'s modular design overcomes this limitation by enabling the integration of new state-of-the-art models as they become available.
Moreover, our experiments in \cref{tab:generalization} indicate that when trained reward models are unavailable, we can use sufficiently strong LLMs in place of trained RMs. For example, we use GPT4o-mini as a reward model for commonsense and logical reasoning. 

\newpage
\onecolumn

\section{Prompt for the Reviewer and the Refiner}
\label{sec:prompt}
\vspace{1em}
\begin{user_example}[frametitle={Reviewer’s Prompt}]
Your task is to provide step-by-step feedback to the current solution.

You will be given a math problem and a current solution, along with the scores for each step based on its correctness.

- You will find (Score: n/10) at the end of each step.

- The maximum (best) score is 10, which means that this step is 100\% correct (and 0\% incorrect).

- The minimum (worst) score is 0, which means that this step is 100\% incorrect (and 0\% correct).

- Pay attention to the steps having scores lower than 6, and carefully identify the errors in those steps.

- Provide your explanation of the error and how it can be fixed. DO NOT propose a new solution, just the explanation.

Question:

\{question\}

Current Solution:

\{solution\}

Let's review the steps with lower scores and identify the errors.

\{feedback\}
\end{user_example}
\vspace{1em}
\begin{user_example}[frametitle={Refiner’s Prompt}]
Your task is to fix the error in the given solution, based on the teacher's feedback.

- After reviewing the solution and feedback, provide a better solution to fix the mistakes you found.

Question:

\{question\}

Current Solution:

\{solution\}

Teacher Feedback:

\{reviewer\_feedback\}

Revised Solution:

Let's review the solution and the feedback, and then refine the current solution.

After reviewing the solution and feedback, let's solve this math problem again.

\{refined\_solution\}
\end{user_example}

\newpage
\vspace{5pt}
\section{Prompt for GPT4o-mini as PRM}
\label{sec:gpt4o_prompt}
\vspace{5pt}
\begin{user_example}[frametitle={GPT4o-mini as PRM Prompt}]
You will be given a question and its corresponding step-by-step solution.

Your task is to evaluate the correctness of each step in the solution by assigning a score between 1 and 10. 

Your response should be formatted as follows:

score: <comma separated scores>

where the length of the list corresponds to the number of steps in the solution. 

Each step is clearly separated in the solution by <Step 1:>, <Step 2:> ... <Step k:>, your output score list should be the exact same length as k.

Question: \{question\}

Solution: \{solution\}

score: 
\end{user_example}

\section{Dataset Statistics and Licenses}
We provide the sample sizes and licenses of the datasets used in this work in \cref{tab:datasets}. All the datasets are in English and all datasets are used in a fashion consistent with their intended use.
\begin{table*}[ht!]
    \centering
    \begin{tabular}{lcc}
    \toprule
    & Sample Size & License \\\midrule
       MMLU-Math \cite{hendryckstest2021} & 974 & MIT License \\
       MATH \cite{hendrycksmath2021} & 500 & MIT License \\
       SVAMP \cite{patel2021nlp} & 1,000 & MIT License \\
       GSM8K \cite{cobbe2021training} & 1,312 & MIT License \\
       SAT \cite{zhong2023agieval} & 220 & MIT License \\
       ARC-Challenge \cite{clark2018think} & 200 & CC-by-SA-4.0\\
       Date Understanding \cite{srivastava2022beyond} & 200 & Apache License\\\bottomrule
    \end{tabular}
    \caption{The statistics and licenses of the datasets used in this study.}
    \label{tab:datasets}
\end{table*}

\newpage
\section{Qualitative Examples of Refinement}
\begin{table*}[!h]
\begin{tabularx}{\textwidth}{X}
\Xhline{2\arrayrulewidth}
\\[-1.5ex]
\textcolor{MidnightBlue}{\textsc{Question}}: Josh decides to try flipping a house. He buys a house for \$80,000 and then puts in \$50,000 in repairs. This increased the value of the house by 150\%. How much profit did he make? (Gold answer: \textbf{70000})\\\\
\textcolor{MidnightBlue}{\textsc{Solver's Output (Before Refinement)}}:\\Step 1: Calculate the value of the house after repairs. The house is worth 80,000 after repairs. It is 150\% higher than before. So the value is (1 + 150/100) * 80,000. 80,000 * (3/2) is 120,000. So the house is worth 120,000 after repairs. (Score: 2/10)\\Step 2: Calculate the cost of repairs. The cost of repairs is 50,000. (Score: 2/10)\\Step 3: Calculate the profit. The profit is the difference between the value after repairs and the cost before repairs. 120,000 - 80,000 - 50,000 is 30,000 (\textcolor{red}{incorrect}).\\\\
\textcolor{MidnightBlue}{\textsc{Reviewer's Output}}:\\
Step 1: The low score of 2 suggests an error in this step. The problem states that the value of the house increased by 150\%, not that the value is 150\% higher than before.\\Step 2: The cost of repairs is indeed \$50,000.\\Step 3: The low score of 3 suggests an error in this step. The profit would be: 200,000 (value after repairs) - 80,000 (cost before repairs) - 50,000 (cost of repairs) = 70,000\\\\
\textcolor{MidnightBlue}{\textsc{Refiner's Output}}:\\
Step 1: Calculate the value of the house after repairs. The house is worth 80,000 after repairs. It is increased by 150\%. So the value is 80,000 x (1 + 150/100) = 80,000 x 2.5 = 200,000.\\Step 2: Calculate the cost of repairs. The cost of repairs is 50,000.\\Step 3: Calculate the profit. The profit is the difference between the value after repairs and the cost before repairs. 200,000 - 80,000 - 50,000 is 70,000 (\textcolor{ForestGreen}{correct}).
\\
\Xhline{2\arrayrulewidth}
\end{tabularx}
\caption{A qualitative example of our multi-agent refinement on GSM8K using \llama~. The score at the end of each step is provided by the PRM.}
\label{tab:qualitative_GSM8K}
\end{table*}

\begin{table*}[!h]
\begin{tabularx}{\textwidth}{X}
\Xhline{2\arrayrulewidth}
\\[-1.5ex]
\textcolor{MidnightBlue}{\textsc{Question}}: The Greek army contained two types of soldiers: the upper class and the lower class soldiers. If there were a total of 5 upper class soldiers, and 10 lower class soldiers in a certain part of Athens, and the battle of Thermopylae demands a force of 4 upper class soldiers and 8 lower class soldiers, how many different battalions can be sent? (Gold answer: \textbf{225})\\\\
\textcolor{MidnightBlue}{\textsc{Solver's Output (Before Refinement)}}:\\Step 1: Find out the number of upper class soldiers left. We have 5 upper class soldiers and we need 4. So we have 5 - 4 = 1 upper class soldiers left. (Score: 9/10)\\Step 2: Find out the number of lower class soldiers left. We have 10 lower class soldiers and we need 8. So we have 10 - 8 = 2 lower class soldiers left. (Score: 9/10)\\Step 3: Find out the number of different battalions. We have 1 upper class soldier and 2 lower class soldiers. The number of different battalions is the number of ways we can arrange 1 upper class soldier and 2 lower class soldiers. This is 3. (Score: 5/10) (\textcolor{red}{incorrect}).\\\\
\textcolor{MidnightBlue}{\textsc{Reviewer's Output}}:\\
Step 1 and Step 2: The scores of 9/10 suggest that these steps are mostly correct. However, we should be careful when calculating the number of upper and lower class soldiers left. \\Step 3: The low score of 5 suggests an error in this step. The problem asks for the number of different battalions, not the number of ways to arrange the soldiers. The correct solution is to find the number of ways to choose 4 upper class soldiers from the available 5 and 8 lower class soldiers from the available 10.\\\\
\textcolor{MidnightBlue}{\textsc{Refiner's Output}}:\\
Step 1: Find out the number of upper class soldiers needed. We need 4 upper class soldiers.\\Step 2: Find out the number of lower class soldiers needed. We need 8 lower class soldiers.\\Step 3: Find out the number of different battalions. We can choose 4 upper class soldiers from the 5 available in 5C4 = 5 ways. We can choose 8 lower class soldiers from the 10 available in 10C8 = 45 ways. The number of different battalions is the number of ways to choose 4 upper class soldiers and 8 lower class soldiers, which is 5 × 45 = 225. (\textcolor{ForestGreen}{correct}).
\\
\Xhline{2\arrayrulewidth}
\end{tabularx}
\caption{A qualitative example of our multi-agent refinement on MATH using \llama~. The score at the end of each step is provided by the PRM.}
\label{tab:qualitative_MATH}
\end{table*}

\clearpage

\section{Qualitative Examples of Over-Correction}
\begin{table*}[!h]
\begin{tabularx}{\textwidth}{X}
\Xhline{2\arrayrulewidth}
\\[-1.5ex]
\textcolor{MidnightBlue}{\textsc{Question}}: A new program had 60 downloads in the first month. The number of downloads in the second month was three times as many as the downloads in the first month, but then reduced by 30\% in the third month. How many downloads did the program have total over the three months? (Gold answer: \textbf{366})\\\\
\textcolor{MidnightBlue}{\textsc{Solver's Output (Before Refinement)}}:\\Step 1: Determine the number of downloads for the second month. The number of downloads for the second month was three times as many as the first month, so it was 60 * 3 = 180. (Score: 10/10)\\
Step 2: Determine the number of downloads for the third month. There was a reduction of 30\%, so we calculate 180 * (1 - 0.30) = 180 * 0.7 = 126. (Score: 10/10)\\
Step 3: Determine the total number of downloads for the three months. There were 60 downloads in the first month, and 180 downloads in the second month. So the total number of downloads for the first two months is 60 + 180 = 240. Then, there were 126 downloads in the third month. So the total number of downloads over the three months is 240 + 126 = 366. (Score: 6/10) (\textcolor{ForestGreen}{correct}).\\\\
\textcolor{MidnightBlue}{\textsc{Reviewer's Output}}:\\
Step 1-2: The high scores of 10 suggest that these steps are correct. 
\\Step 3: The score of 6 suggests that this step is mostly correct. However, the improvement you could make is to explicitly state the calculation for the total number of downloads, rather than just stating the result.\\\\
\textcolor{MidnightBlue}{\textsc{Refiner's Output}}:\\
Step 1: Determine the downloads in the second month. Three times 60 is 3 x 60 = 180. Then 30\% of 180 is 180 x 0.3 = 54. 180 - 54 is 126. So 126 downloads in the second month.\\
Step 2: Determine the total downloads over the three months. 60 downloads in the first month, and 126 downloads in the second month, and 126 - 54 = 72 downloads in the third month. 60 + 126 + 72 is 258. The answer is 258. (\textcolor{red}{incorrect}).
\\
\Xhline{2\arrayrulewidth}
\end{tabularx}
\caption{A qualitative example of over-correction on GSM8K using \llama.}
\label{tab:overcorrect_GSM8K}
\end{table*}

\begin{table*}[!h]
\begin{tabularx}{\textwidth}{X}
\Xhline{2\arrayrulewidth}
\\[-1.5ex]
\textcolor{MidnightBlue}{\textsc{Question}}: If $f(x) = \frac{3x-2}{x-2}$, what is the value of $f(-2) +f(-1)+f(0)$? Express your answer as a common fraction. (Gold answer: \textbf{$\frac{14}{3}$})\\\\
\textcolor{MidnightBlue}{\textsc{Solver's Output (Before Refinement)}}:\\Step 1: Determine the values of $f(-2)$, $f(-1)$, and $f(0)$ by plugging the values into the expression for $f(x)$. \\
$f(-2) = \frac{3(-2)-2}{-2-2} = \frac{-6-2}{-4} = \frac{-8}{-4} = 2$,\\\\ $f(-1) = \frac{3(-1)-2}{-1-2} = \frac{-3-2}{-3} = \frac{-5}{-3} = \frac{5}{3}$,\\\\ $f(0) = \frac{3(0)-2}{0-2} = \frac{0-2}{-2} = \frac{-2}{-2} = 1$. (Score: 10/10)\\
Step 2: Add the values of $f(-2)$, $f(-1)$, and $f(0)$ to obtain $f(-2) + f(-1) + f(0) = 2+\frac{5}{3} + 1$\\\\ $\frac{6}{3} + \frac{5}{3} + \frac{3}{3} = \frac{14}{3}$. \\
Final Answer: The final answer is $\boxed{\frac{14}{3}}$. I hope it is correct.  (Score: 7/10) (\textcolor{ForestGreen}{correct}).\\\\
\textcolor{MidnightBlue}{\textsc{Reviewer's Output}}:\\
Step 1: The scores of 10/10 suggest that this step is correct.\\Step 2: The high score of 7 suggests that this step is mostly correct. However, it might be beneficial to include intermediate steps for clarity.\\\\
\textcolor{MidnightBlue}{\textsc{Refiner's Output}}:\\
Step 1: We can calculate $f(-2)$, $f(-1)$, and $f(0)$ using the given expression for $f(x)$:
$f(-2) = \frac{3(-2)-2}{-2-2} = \frac{-4}{-4}=1$,\\\\$f(-1) = \frac{3(-1)-2}{-1-2} = \frac{-3}{-3}=1$,\\\\$f(0) = \frac{3(0)-2}{0-2} = \frac{-2}{-2} = 1$.\\
Step 2: Now we can add these values together:
$f(-2) + f(-1) + f(0) = 1 + 1 + 1 = \frac{3}{1}$\\
Final Answer: The final answer is $\boxed{\frac{3}{1}}$. I hope it is correct. (\textcolor{red}{incorrect}).
\\
\Xhline{2\arrayrulewidth}
\end{tabularx}
\caption{A qualitative example of over-correction on MATH using \llama~.}
\label{tab:overcorrect_MATH}
\end{table*}